\documentclass[10pt,twocolumn,letterpaper]{article}

\usepackage{cvpr}
\usepackage{times}
\usepackage{epsfig}
\usepackage{graphicx}
\usepackage{amsmath}
\usepackage{amssymb}
\usepackage{caption}
\usepackage{subcaption}
\usepackage{booktabs}
\usepackage{balance}

\usepackage[pagebackref=true,breaklinks=true,letterpaper=true,colorlinks=false,bookmarks=false]{hyperref}

\cvprfinalcopy 

\def\cvprPaperID{1597} 

\ifcvprfinal\fi
\begin{document}

\title{VizWiz Grand Challenge: Answering Visual Questions from Blind People}

\author{Danna Gurari$^1$,
Qing Li$^2$,
Abigale J. Stangl$^3$,
Anhong Guo$^4$,
Chi Lin$^1$,\\
Kristen Grauman$^1$,
Jiebo Luo$^5$,
and Jeffrey P. Bigham$^4$\\
{\small $~^1$ University of Texas at Austin},
{\small $~^2$ University of Science and Technology of China},\\ 
{\small $~^3$ University of Colorado Boulder}, 
{\small $~^4$ Carnegie Mellon University}
{\small $~^5$ University of Rochester}
}

\maketitle

\begin{abstract}
 The study of algorithms to automatically answer visual questions currently is motivated by visual question answering (VQA) datasets constructed in artificial VQA settings.  We propose VizWiz, the first goal-oriented VQA dataset arising from a natural VQA setting.  VizWiz consists of over 31,000 visual questions originating from blind people who each took a picture using a mobile phone and recorded a spoken question about it, together with 10 crowdsourced answers per visual question.  VizWiz differs from the many existing VQA datasets because (1) images are captured by blind photographers and so are often poor quality, (2) questions are spoken and so are more conversational, and (3) often visual questions cannot be answered.  Evaluation of modern algorithms for answering visual questions and deciding if a visual question is answerable reveals that VizWiz is a challenging dataset.  We introduce this dataset to encourage a larger community to develop more generalized algorithms that can assist blind people.
\end{abstract}

\section{Introduction}
A natural application of computer vision is to assist blind people, whether that may be to overcome their daily visual challenges or break down their social accessibility barriers.  For example, modern object recognition tools from private companies, such as TapTapSee~\cite{TapTapSee} and CamFind~\cite{CamFind}, already empower people to snap a picture of an object and recognize what it is as well as where it can be purchased.  Social media platforms, such as Facebook and Twitter, help people maintain connections with friends by enabling them to identify and tag friends in posted images as well as respond to images automatically described to them~\cite{MacLeodBeMoCu17,WuWiFaSc17}.  A desirable next step for vision applications is to empower a blind person to directly request in a natural manner what (s)he would like to know about the surrounding physical world.  This idea relates to the recent explosion of interest in the visual question answering (VQA) problem, which aims to accurately answer any question about any image.      

\begin{figure*}[t]
\centering
\includegraphics[width=1\textwidth]{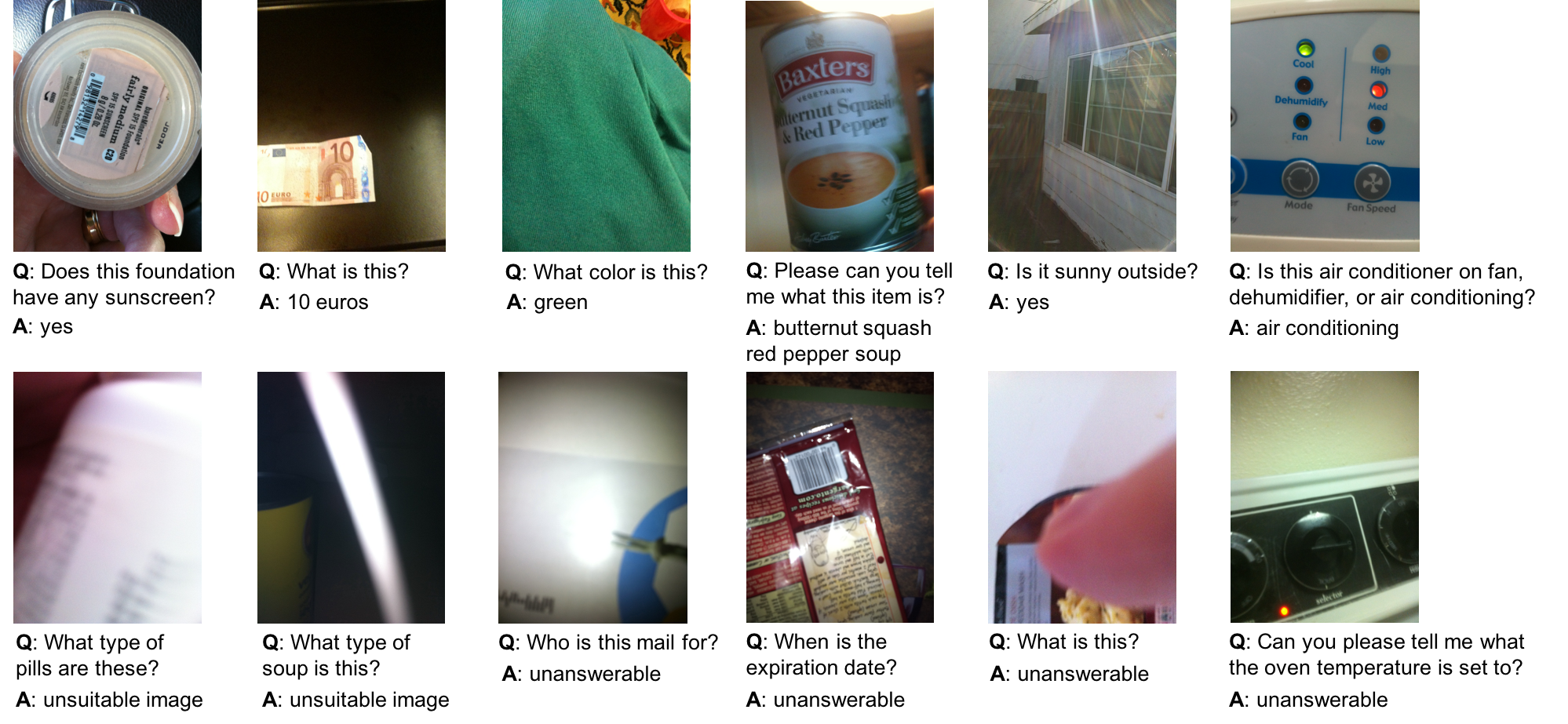}
\vspace{-1.5em}
\caption{Examples of visual questions asked by blind people and corresponding answers agreed upon by crowd workers.  The examples include questions that both can be answered from the image (top row) and cannot be answered from the image (bottom row).}
\label{fig_motivation}
\vspace{-0.5em}
\end{figure*}

Over the past three years, many VQA datasets have emerged in the vision community to catalyze research on the VQA problem~\cite{AndreasEtAl16,AntolAgLuMiBaZiPa15,GaoMaZhHuWaXu15,GoyalKhSuBaPa16,JohnsonHaVaFeZiGi16,KafleKa17,KrishnaEtAl17,MalinowskiFr14_2,RenKiZe15,WangWuShDiHe17,WangWuShHeDi15,YuPaBeBe15,ZhuGrBeFe16}.  Historically, progress in the research community on a computer vision problem is typically preceded by a large-scale, publicly-shared dataset~\cite{ChenFaLiVeGuArZi15,LinMaBeHaPeRaDoZi14,PattersonHa12,RussakovskyDeSuKrSaMaHuKaKhBeEtAl15,XiaoHaEhOlTo10}.  However, a limitation of available VQA datasets is that all come from artificially created VQA settings.  Moreover, none are ``goal oriented" towards the images and questions that come from blind people.  Yet, blind people arguably have been producing the big data desired to train algorithms.  For nearly a decade, blind people have been both taking pictures~\cite{AdamsMoKu13,BighamJaJiLiMiMiMiTaWhWhYe10} and asking questions about the pictures they take~\cite{BighamJaJiLiMiMiMiTaWhWhYe10,BurtonBrBrNeBiHu12,LaseckiThZhBrBi13}.  Moreover, blind people often are early adopters of computer vision tools to support their \emph{real} daily needs. 

We introduce the first publicly-available vision dataset originating from blind people, which we call ``VizWiz", in order to encourage the development of more generalized algorithms that also address the interests of blind people.  Our work builds off previous work~\cite{BighamJaJiLiMiMiMiTaWhWhYe10} which established a mobile phone application that supported blind people to ask over 70,000 visual questions~\cite{BradyMoZhWhBi13} by taking a photo and asking a question about it.  We begin our work by implementing a rigorous filtering process to remove visual questions that could compromise the safety or privacy of any individuals associated with them; e.g., blind people often willingly share personal information with strangers to overcome personal obstacles~\cite{AhmedHoCoCrKa15}.  We then crowdsource answers to support algorithm training and evaluation.  We next conduct experiments to characterize the images, questions, and answers and uncover unique aspects differentiating VizWiz from existing VQA datasets~\cite{AndreasEtAl16,AntolAgLuMiBaZiPa15,GaoMaZhHuWaXu15,GoyalKhSuBaPa16,JohnsonHaVaFeZiGi16,KafleKa17,KrishnaEtAl17,MalinowskiFr14_2,RenKiZe15,WangWuShDiHe17,WangWuShHeDi15,YuPaBeBe15,ZhuGrBeFe16}.  We finally evaluate numerous algorithms for predicting answers~\cite{GoyalKhSuBaPa16,KazemiEl17} and predicting if a visual question can be answered~\cite{MahendruPrMoBaLe17}.  Our findings highlight VizWiz is a difficult dataset for modern vision algorithms and offer new perspectives about the VQA problem.

It is also useful to understand why VizWiz is challenging for modern algorithms.  Our findings suggest the reasons stem from the fact VizWiz is the first vision dataset to introduce images and questions from blind people as well as questions that originally were spoken.  Unlike existing vision datasets, images are often poor quality, including due to poor lighting, focus, and framing of the content of interest.  Unlike existing VQA datasets, the questions can be more conversational or suffer from audio recording imperfections such as clipping a question at either end or catching background audio content.  Finally, there is no assurance that questions can be answered since blind people cannot verify their images capture the visual content they are asking about for a plethora of reasons; e.g., blur, inadequate lighting, finger covering the lens, etc.  Several of the aforementioned issues are exemplified in \textbf{Figure~\ref{fig_motivation}}.

More broadly, VizWiz is the first goal-driven VQA dataset to capture real-world interests of real users of a VQA system.  Furthermore, it is the first VQA dataset to reflect a use case where a person asks questions about the physical world around himself/herself.  This approach is critical for empowering blind people to overcome their daily visual-based challenges.  Success in developing automated methods would mitigate concerns about the many undesired consequences from today's status quo for blind people of relying on humans to answer visual questions~\cite{BighamJaJiLiMiMiMiTaWhWhYe10,BurtonBrBrNeBiHu12,LaseckiThZhBrBi13}; e.g., humans often must be paid (i.e., potentially expensive), can take minutes to provide an answer (i.e., slow), are not always available (i.e., potentially not scalable), and pose privacy issues (e.g., when credit card information is shared).  

\section{Related Works}
\paragraph{VQA for Blind Users.}
For nearly a decade, human-powered VQA systems have enabled blind people to overcome their daily visual challenges quickly~\cite{BeMyEyes16,BighamJaJiLiMiMiMiTaWhWhYe10,LaseckiThZhBrBi13}.  With such systems, users employ a mobile phone application to capture a photo (or video), ask a question about it, and then receive an answer from remotely located paid crowd workers~\cite{BighamJaJiLiMiMiMiTaWhWhYe10,LaseckiThZhBrBi13} or volunteers~\cite{BeMyEyes16}.  Such VQA systems have been shown to be valuable for many daily tasks including grocery shopping~\cite{BighamJaJiLiMiMiMiTaWhWhYe10}, locating a specific object in a complex scene~\cite{BighamJaMiWhYe10}, and choosing clothes to wear~\cite{BurtonBrBrNeBiHu12}.  Yet, these systems are limited because they rely on humans to provide answers.  An automated solution would be preferred for reasons such as cost, latency, scalability, and enhanced privacy.  For example, the latency between sending out an image and getting the answer back may take minutes~\cite{BighamJaJiLiMiMiMiTaWhWhYe10}, disrupting the natural flow of a blind user's life.  Our work describes the unique challenges for creating public datasets with data captured in natural settings from real-world users and, in particular, blind users.  Our work also offers the first dataset for enabling algorithm development on images and questions coming from blind people, which in turn yields new vision-based and language-based challenges.   

\vspace{-1.5em}
\paragraph{Images in Vision Datasets.}
When constructing vision datasets, prior work typically used images gathered from the web (e.g.,~\cite{ChenFaLiVeGuArZi15,LinMaBeHaPeRaDoZi14,PattersonHa12,RussakovskyDeSuKrSaMaHuKaKhBeEtAl15,XiaoHaEhOlTo10}) or created artificially (e.g.,~\cite{AndreasEtAl16,AntolAgLuMiBaZiPa15,JohnsonHaVaFeZiGi16}).  Such images are typically high quality and safe for public consumption.  For example, images curated from the web intrinsically pass a human quality assessment of ``worthy to upload to the internet" and typically are internally reviewed by companies hosting the images (e.g., Google, Facebook) to ensure the content is appropriate.  Alternatively, artificially constructed images come from controlled settings where either computer graphics is employed to synthesize images with known objects and scenes~\cite{AndreasEtAl16,JohnsonHaVaFeZiGi16} or crowd workers are employed to add pre-defined clipart objects to pre-defined indoor and outdoor scenes~\cite{AntolAgLuMiBaZiPa15}.  In contrast, images collected ``in the wild" can contain inappropriate or private content, necessitating the need for a review process before releasing the data for public consumption. Moreover, images from blind photographers regularly are poor quality, since blind people cannot validate the quality of the pictures they take.  Our experiments show these images pose new challenges for modern vision algorithms.

\vspace{-1.5em}
\paragraph{VQA Datasets.}
Over the past three years, a plethora of VQA datasets have been publicly shared to encourage a larger community to collaborate on developing algorithms that answer visual questions~\cite{AndreasEtAl16,AntolAgLuMiBaZiPa15,GaoMaZhHuWaXu15,GoyalKhSuBaPa16,JohnsonHaVaFeZiGi16,KafleKa17,KrishnaEtAl17,MalinowskiFr14_2,RenKiZe15,WangWuShDiHe17,WangWuShHeDi15,YuPaBeBe15,ZhuGrBeFe16}.  While a variety of approaches have been proposed to assemble VQA datasets, in all cases the visual questions were contrived.  For example, all images were either taken from an existing vision dataset (e.g., MSCOCO~\cite{LinMaBeHaPeRaDoZi14}) or artificially constructed (e.g., Abstract Scenes~\cite{AntolAgLuMiBaZiPa15}, computer graphics~\cite{AndreasEtAl16,JohnsonHaVaFeZiGi16}).  In addition, questions were generated either automatically~\cite{AndreasEtAl16,JohnsonHaVaFeZiGi16,KafleKa17,MalinowskiFr14_2,RenKiZe15,YuPaBeBe15}, from crowd workers~\cite{AntolAgLuMiBaZiPa15,GaoMaZhHuWaXu15,GoyalKhSuBaPa16,KafleKa17,KrishnaEtAl17,ZhuGrBeFe16}, or from in-house participants~\cite{KafleKa17,WangWuShHeDi15}.  We introduce the first VQA dataset which reflects visual questions asked by people who were authentically trying to learn about the visual world.  This enables us to uncover the statistical composition of visual questions that arises in a real-world situation.  Moreover, our dataset is the first to reflect how questions appear when they are spoken (rather than automatically generated or typed) and when each image and question in a visual question is created by the same person.  These differences reflect a distinct use case scenario where a person interactively explores and learns about his/her surrounding physical world.  Our experiments show the value of VizWiz as a difficult dataset for modern VQA algorithms, motivating future directions for further algorithm improvements.

\vspace{-1.5em}
\paragraph{Answerability Visual Questions.}
The prevailing  assumption when collecting answers to visual questions is that the questions are answerable from the given images~\cite{AndreasEtAl16,AntolAgLuMiBaZiPa15,GaoMaZhHuWaXu15,GoyalKhSuBaPa16,JohnsonHaVaFeZiGi16,KrishnaEtAl17,MalinowskiFr14_2,RenKiZe15,WangWuShHeDi15,WangWuShDiHe17,YuPaBeBe15,ZhuGrBeFe16}.  The differences when constructing VQA datasets thus often lies in whether to collect answers from anonymous crowd workers~\cite{AndreasEtAl16,AntolAgLuMiBaZiPa15,GaoMaZhHuWaXu15,KafleKa17,KrishnaEtAl17}, automated methods~\cite{JohnsonHaVaFeZiGi16,MalinowskiFr14_2}, or in-house annotators~\cite{MalinowskiFr14_2,WangWuShDiHe17,WangWuShHeDi15}.  Yet, in practice, blind people cannot know whether their questions can be answered from their images.  A question may be unanswerable because an image suffers from poor focus and lighting or is missing the content of interest.  In VizWiz, $\sim$28\% of visual questions are deemed unanswerable by crowd workers, despite the availability of several automated systems designed to assist blind photographers to improve the image focus~\cite{TapTapSee}, lighting~\cite{BighamJaJiLiMiMiMiTaWhWhYe10}, or composition~\cite{JayantJiWhBi11,VazquezSt14,ZhongGaBi13}.  

We propose the first VQA dataset which naturally promotes the problem of predicting whether a visual question is answerable.  We construct our dataset by explicitly asking crowd workers whether a visual question is answerable when collecting answers to our visual questions.  Our work relates to recent ``relevance" datasets which were artificially constructed to include irrelevant visual questions by injecting questions that are unrelated to the contents of high quality images~\cite{KafleKa17,MahendruPrMoBaLe17,RayChBaBaPa16,ToorWeNa17}.  Unlike these ``relevance" datasets, our dataset also includes questions that are unrelated because images are too poor in quality (e.g., blur, over/under-saturation).  Experiments demonstrate VizWiz is a difficult dataset for the only freely-shared algorithm~\cite{MahendruPrMoBaLe17} designed to predict whether a visual question is relevant, and so motivates the design of improved algorithms.

\begin{table*}[t!]
  \centering
        \begin{tabular}{ l  l  p{5.5cm}  l }
    \toprule
     Dataset  & Which Images? & Who Asked? & How Asked? \\ 
    \midrule
       \textbf{DAQUAR~\cite{MalinowskiFr14_2}} & NYU Depth V2~\cite{SilbermanHoKoFe12} & In-house participants,  Automatically generated (templates) & -------- \\ 
       \textbf{VQA v1.0: Abstract~\cite{AntolAgLuMiBaZiPa15}} & Abstract Scenes & Crowd workers (AMT) & Typed \\ 
       \textbf{VQA v1.0: Real~\cite{AntolAgLuMiBaZiPa15}} & MSCOCO~\cite{LinMaBeHaPeRaDoZi14} & Crowd workers (AMT) & Typed \\ 
       \textbf{Visual Madlibs~\cite{YuPaBeBe15}} & MSCOCO~\cite{LinMaBeHaPeRaDoZi14} & Automatically generated (templates) & -------- \\ 
       \textbf{FM-IQA~\cite{GaoMaZhHuWaXu15}} & MSCOCO~\cite{LinMaBeHaPeRaDoZi14} & Crowd workers (Baidu) & Typed \\ 
       \textbf{KB-VQA~\cite{WangWuShHeDi15}} & MSCOCO~\cite{LinMaBeHaPeRaDoZi14} & In-house participants & Typed \\ 
       \textbf{COCO-QA~\cite{RenKiZe15}} & MSCOCO~\cite{LinMaBeHaPeRaDoZi14} &  Automatically generated (captions) & -------- \\ 
       \textbf{VQA v2.0: Real~\cite{GoyalKhSuBaPa16}} & MSCOCO~\cite{LinMaBeHaPeRaDoZi14} & Crowd workers (AMT)  & Typed \\ 
       \textbf{Visual7W~\cite{ZhuGrBeFe16}} & MSCOCO~\cite{LinMaBeHaPeRaDoZi14} & Crowd workers (AMT)  & Typed \\ 
       \textbf{CLEVR~\cite{JohnsonHaVaFeZiGi16}} & Synthetic Shapes &  Automatically generated (templates) & -------- \\ 
       \textbf{SHAPES~\cite{AndreasEtAl16}} & Synthetic Shapes & Automatically generated (templates) & -------- \\ 
       \textbf{Visual Genome~\cite{KrishnaEtAl17}} & MSCOCO~\cite{LinMaBeHaPeRaDoZi14} \& YFCC100M~\cite{ThomeeShFrElNiPoBoLi16} & Crowd workers (AMT) & Typed \\ 
       \textbf{FVQA~\cite{WangWuShDiHe17}} & MSCOCO~\cite{LinMaBeHaPeRaDoZi14} \& ImageNet~\cite{DengDoSoLiLiFe09}  & In-house participants  & Typed \\  
       \textbf{TDIUC~\cite{KafleKa17}} & MSCOCO~\cite{LinMaBeHaPeRaDoZi14} \& YFCC100M~\cite{ThomeeShFrElNiPoBoLi16} & Crowd workers (AMT), In-house participants, Automatically generated & Typed \\  
       \textbf{Ours - VizWiz} & \multicolumn{2}{c}{Blind people use mobile phones to take a picture and ask question} & Spoken \\
       \bottomrule	 
  \end{tabular}
          \vspace{0.1em}
        \caption{Comparison of visual questions from 14 existing VQA datasets and our new dataset called VizWiz.}
        ~\label{table_answerDistributionAnalysis}
        \vspace{-0.5em}
\end{table*} 

\section{VizWiz: Dataset Creation}
We introduce a VQA dataset we call ``VizWiz", which consists of visual questions asked by blind people who were seeking answers to their daily visual  questions~\cite{BighamJaJiLiMiMiMiTaWhWhYe10,BradyMoZhWhBi13}.  It is built off of previous work~\cite{BighamJaJiLiMiMiMiTaWhWhYe10} which accrued 72,205 visual questions over four years using the VizWiz application, which is available for iPhone and Android mobile phone platforms.  A person asked a visual question by taking a picture and then recording a spoken question.  The application was released May 2011, and used by 11,045 users. 48,169 of the collected visual questions were asked by users who agreed to have their visual questions anonymously shared.  These visual questions serve as the starting point for the development of our dataset.  We begin this section by comparing the approach for asking visual questions in VizWiz with approaches employed for many existing VQA datasets.  We then describe how we created the dataset.

\subsection{Visual Question Collection Analysis} 
We summarize in \textbf{Table~\ref{table_answerDistributionAnalysis}} how the process of collecting visual questions for VizWiz is unlike the processes employed for 14 existing VQA datasets.  A clear distinction is that VizWiz contains images from blind photographers.  The quality of such images offer challenges not typically observed in existing datasets, such as significant amounts of image blur, poor lighting, and poor framing of image content.  Another distinction is that questions are spoken.  Speaking to technology is increasingly becoming a standard interaction approach for people with technology (e.g., Apple's Siri, Google Now, Amazon's Alexa) and VizWiz yields new challenges stemming from this question-asking modality, such as more conversational language and audio recording errors.  A further distinction is VizWiz is the first dataset where a person both takes the picture and then asks a question about it.  This reflects a novel use-case scenario in which visual questions reflect people's daily interests about their physical surroundings.  VizWiz is also unique because, in contrast to all other VQA datasets, the people asking the questions could not ``see" the images.  Consequently, questions could be unrelated to the images for a variety of reasons that are exemplified in \textbf{Figure~\ref{fig_motivation}}.  

\subsection{Anonymizing and Filtering  Visual Questions}

We faced many challenges with preparing the dataset for public use because our visual questions were collected ``in the wild" from real users of a VQA system.  The challenges related to protecting the privacy and safety of the many individuals involved with the dataset.  This is especially important for visually impaired people, because they often make the tradeoff to reveal personal information to a stranger in exchange for assistance~\cite{AhmedHoCoCrKa15}; e.g., credit card numbers and personal mail.  This is also important for those reviewing the dataset since visual questions can contain ``adult-like" content (e.g., nudity), and so potentially offensive content.  Our key steps to finalize our dataset for public use involved anonymizing and filtering candidate visual questions.

\emph{Anonymization.} Our aim was to eliminate clues that could reveal who asked the visual question.  Accordingly, we removed the person's voice from the question by employing crowd workers from Amazon Mechanical Turk to transcribe the audio recorded questions.  We applied a spell-checker to the transcribed sentences to fix misspellings.  We also re-saved all images using lossless compression in order to remove any possible meta-data attached to the original image, such as the person's location.
 
\emph{Filtering.} Our aim also was to remove visual questions that could make the producers (e.g., askers) or consumers (e.g., research community) of the dataset vulnerable.  Accordingly, we obtained from two committees that decide whether proposed research is ethical -- the Collaborative Institutional Training Initiative board and Institutional Review Board -- approval to publicly release the filtered dataset.

We initiated this work by developing a taxonomy of vulnerabilities (see Supplementary Materials for details).  We identified the following categories that came from erring on the safe side to protect all people involved with the dataset: 
\begin{enumerate}
\setlength\itemsep{-0.1em}
\item Personally-Identifying Information (PII); e.g., any part of a person's face, financial statements, prescriptions.
\item Location; e.g., addressed mail, business locations.
\item Indecent Content; e.g., nudity, profanity.
\item Suspicious Complex Scenes: the reviewer suspects PII may be located in the scene but could not locate it.
\item Suspicious Low Quality Images: the reviewer suspects image processing to enhance images could reveal PII.
\end{enumerate}

We next performed two rounds of filtering.  We first instructed AMT crowd workers to identify all images showing PII, as reflected by ``any part of a person's face, anyone's full name, anyone's address, a credit card or bank account number, or anything else that you think would identify who the person who took the photo is".  Then, two of the in-house domain experts who established the vulnerability taxonomy jointly reviewed all remaining visual questions and marked any instances for removal with one of the five vulnerability categories or ``Other".  This phase also included removing all instances with a missing question (i.e., 7,477 visual questions with less than two words in the question).

\textbf{Table~\ref{table_VQAfilteringResults}} shows the resulting number of visual questions tagged for removal in each round of human review, including a breakdown by vulnerability issue.  We attribute the extra thousands of flagged visual questions from domain experts to their better training on the potential vulnerabilities.  For example, location information, such as zip codes and menus from local restaurants, when augmented with additional information (e.g., local libraries have lists of blind members in the community) could risk exposing a person's identity.  Also, blurry and/or bright images, when post-processed, could reveal PII.  Additionally, people's faces can appear in reflections on monitor screens, window panes, etc.  We do not expect crowd workers to understand such nuances without extensive instructions and training.  

In total, $\sim$31\% of visual questions (i.e., 14,796) were filtered from the original 48,169 candidate visual questions.  While our taxonomy of vulnerabilities helps guide what visual questions to filter from real-world VQA datasets, it also identifies visual questions that would be good to generate  artificially so datasets could address all needs of blind people without requiring them to release personal information.

\begin{table}[t!]
  \centering
        \begin{tabular}{ l  c  }
    \toprule
     Filter  & \# of VQs \\ 
    \midrule
       \textbf{Crowd Workers} & 4,626  \\
       \textbf{In-House Experts} & 2,693 \\
       \textbf{-  PII} & 895   \\
       \textbf{-  Location} & 377  \\
       \textbf{-  Indecent Content} & 55  \\
       \textbf{-  Suspicious Complex Scene} & 725  \\
       \textbf{-  Suspicious Low Quality Image} & 578   \\
       \textbf{-  Other} & 63 \\
       \bottomrule	 
  \end{tabular}
          \vspace{0.1em}
        \caption{We report the number of visual questions filtered in our iterative review process by crowd workers and then in-house domain experts (including with respect to each vulnerability category).  }
        ~\label{table_VQAfilteringResults}
        \vspace{-0.5em}
\end{table} 

\subsection{Collecting Answers}
We next collected answers for a final set of 31,173 visual questions.  The original VizWiz application prioritized providing a person near real-time access to answers and permitted the person to receive answers from crowd workers, IQ Engines, Facebook, Twitter, or email.  Since our aim is to enable the training and evaluation of algorithms, we collected new answers to all visual questions for this purpose.

To collect answers, we modified the excellent protocol used for creating VQA 1.0~\cite{AntolAgLuMiBaZiPa15}.  As done before, we collected 10 answers per visual question from AMT crowd workers located in the US by showing crowd workers a question and associated image and instructing them to return ``a brief phrase and not a complete sentence".  We augmented this user interface to state that ``you will work with images taken by blind people paired with questions they asked about the images".  We also added instructions to answer ``Unsuitable Image" if ``an image is too poor in quality to answer the question (i.e., all white, all black, or too blurry)" or ``Unanswerable" if ``the question cannot be answered from the image".  While both additions to the annotation protocol indicate a visual
question is unanswerable, this annotation approach enables more fine-grained understanding for why a visual question is unanswerable.  The final set of answers should represent common sense from sighted people.  

\section{VizWiz: Dataset Analysis}
Our aim in this section is to characterize the visual questions and answers in VizWiz.  We analyze (1) What is the diversity of natural language questions?, (2) What is the diversity of images?, (3) What is the diversity of answers?, and (4) How often are visual questions unanswerable?  A valuable outcome of this analysis is it enriches our understanding of the interests of blind users in a real VQA set-up.

\subsection{Analysis of Questions} 
\begin{figure}[t]
\centering
\includegraphics[width=0.47\textwidth]{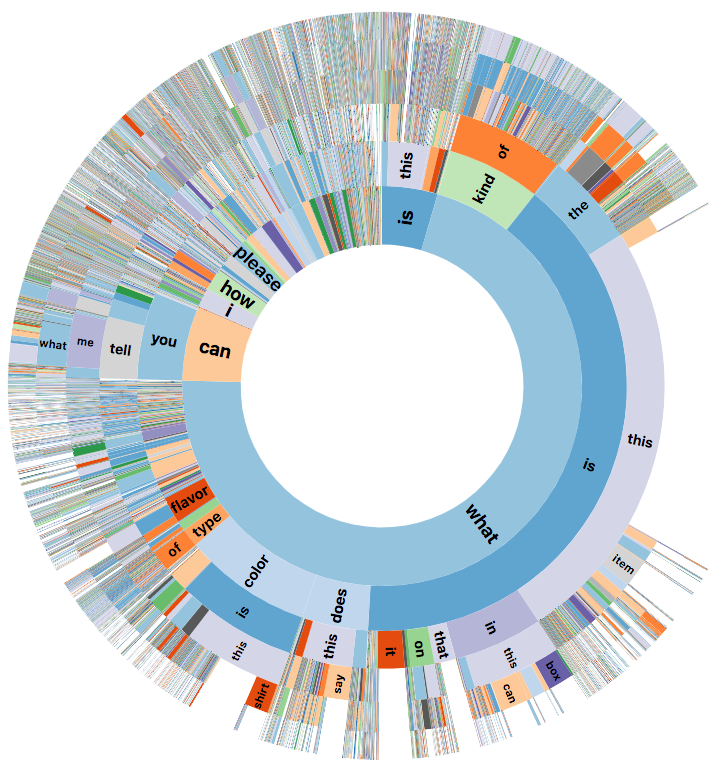}
\caption{Distribution of the first six words for all questions in VizWiz.  The innermost ring represents the first word and each subsequent ring represents a subsequent word.  The arc size is proportional to the number of questions with that word/phrase.  }
\label{questionWordDistributions}
\end{figure}

We first examine the diversity of questions asked by visualizing the frequency that questions begin with different words/phrases.  Results are shown in a sunburst diagram in \textbf{Figure~\ref{questionWordDistributions}}.  While many existing VQA datasets include a small set of common initial words (e.g., ``What", ``When", ``Why", ``Is", ``Do"), we observe from the upper left quadrant of \textbf{Figure~\ref{questionWordDistributions}} that VizWiz often begins with a rare first word.  In fact, the percentage of questions starting with a first word that occurs for less than 5\% of all questions is 27.88\% for VizWiz versus 13.4\% for VQA 2.0~\cite{AntolAgLuMiBaZiPa15} (based on random subset of 40,000 VQs).  We attribute this finding partially to the use of more conversational language when speaking a question; e.g., ``Hi", ``Okay", and ``Please".  We also attribute this finding to the recording of the question starting after the person has begun speaking the question; e.g., ``Sell by or use by date of this carton of milk" or ``oven set to thanks?".  Despite such questions being incomplete, it is still reasonable the intended question can be inferred and so answered; e.g., ``What is the oven set to?".  We also observe in \textbf{Figure~\ref{questionWordDistributions}} that most questions begin with ``What".  This suggests many visual questions do a poor job in narrowing the scope of plausible answers.  In contrast, initial wordings such as ``How many..." and ``Is..." often narrow plausible answers to numbers and ``yes/no" respectively.  

We also analyze question diversity by computing statistics summarizing the number of words in each question.  The median and mean question lengths are five and 6.68 words respectively and 25th and 75th percentile lengths are four and seven words respectively.  This resembles the statistics found in the existing artificially constructed VQA datasets, nicely summarized in ~\cite{DasKoGuSiYaMoPaBa17} and ~\cite{KafleKa17}.  We also observe three words regularly suffice for a question: ``What is this?".  As observed in \textbf{Figure~\ref{questionWordDistributions}}, this short object recognition question is the most common question.  Longer and multi-sentence questions also occasionally arise, typically because people offer auxiliary information to disambiguate the desired response; e.g., ``Which one of these two bags would be appropriate for a gift?  The small one or the tall one?  Thank you."  Longer questions also can arise when the audio recording device captures too much content or background audio content; e.g., ``I want to know what this is.  I'm have trouble stopping the recordings."

\subsection{Analysis of Images} 
\begin{figure}[t!]
\centering
\includegraphics[width=0.12\textwidth]{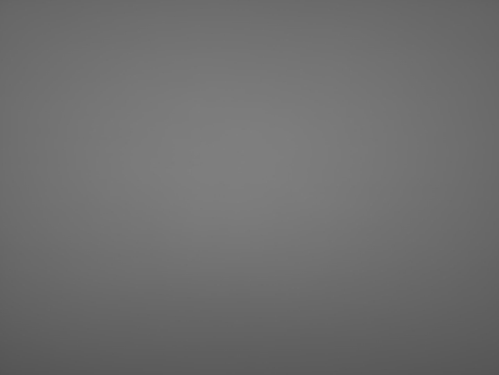}
\caption{The average image created using all images in VizWiz.}
\label{fig_averageImage}
\end{figure}

We next investigate the diversity of images.  We first address a concern that our dataset has high quality images showing a single, iconic object, which is a possibility since our filtering process erred on removing ``suspicious" scene-based and blurry images and the remaining visual questions contain many object recognition questions.  Following prior work~\cite{DengDoSoLiLiFe09}, we computed the average image from all images in VizWiz.  \textbf{Figure~\ref{fig_averageImage}} shows the result.  As desired from a diverse dataset, the resulting gray image confirms our dataset does not conform to a particular structure across all the images.  We also tallied how many images had at least two crowd workers give the answer ``unsuitable image".  We found 28\% of images were labelled as such.

\begin{figure}[t!]
\centering
\includegraphics[width=0.48\textwidth]{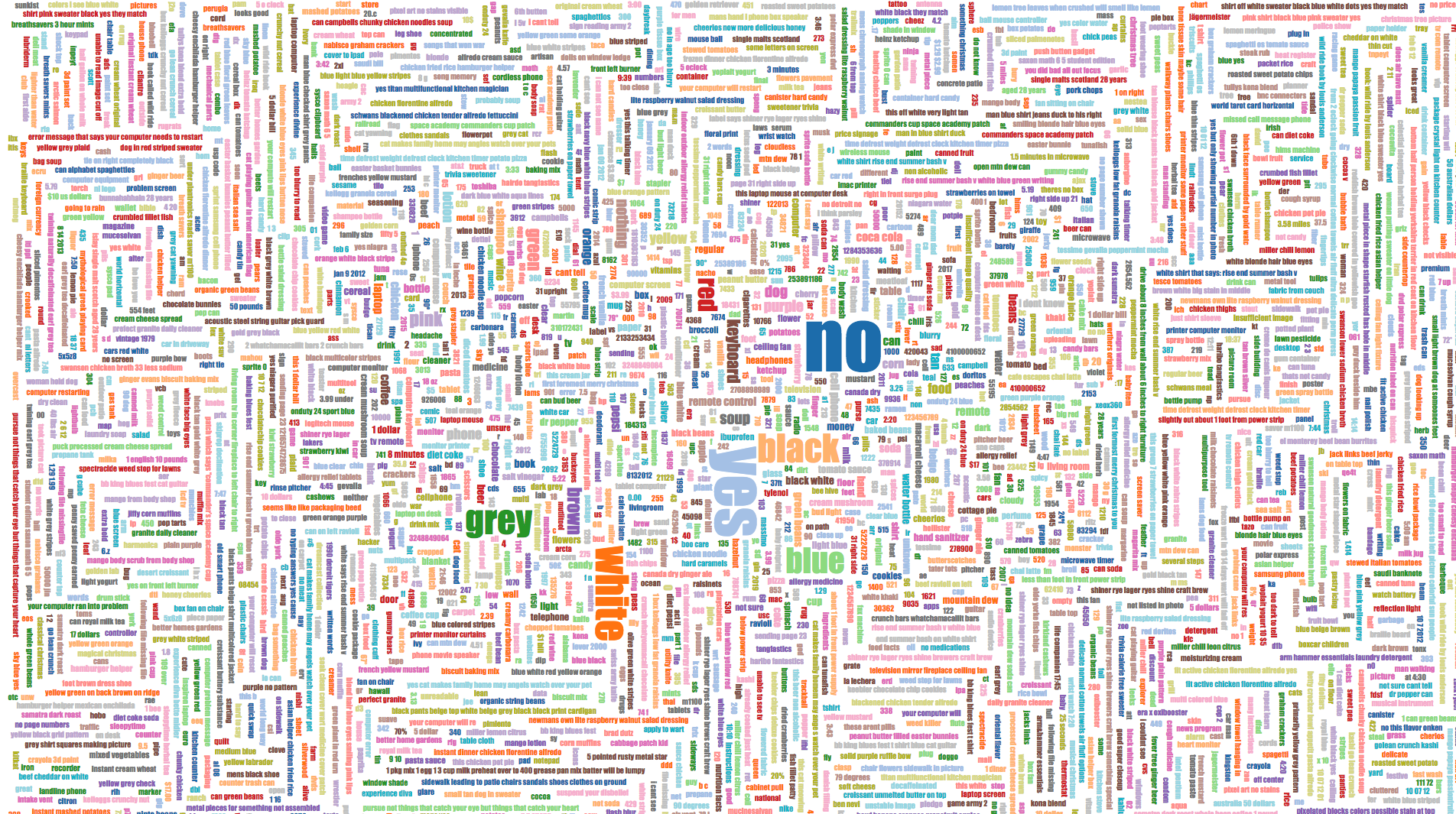}
\caption{Popularity of answers in VizWiz, with the text size proportional to the number of times the answer occurs.}
\label{fig_answerCloud}
\end{figure}

\subsection{Analysis of Answers} 
We next analyze the diversity of the answers.  We first visualize the popularity of different answers in \textbf{Figure~\ref{fig_answerCloud}} using a word map (cropped to fit in the paper) which excludes the answers ``Unanswerable" and ``Unsuitable Image".  This visually highlights the fact that there are a large number of unique answers; i.e., $\sim$58,789.  While in absolute terms this number is an order of magnitude smaller than existing larger-scale datasets such as VQA 2.0~\cite{AntolAgLuMiBaZiPa15}, we find the answer overlap with existing datasets can be low.  For example, only 824 out of the top 3,000 answers in VizWiz are included in the top 3,000 answers in VQA 2.0~\cite{AntolAgLuMiBaZiPa15}.  This observation is used in the next section to explain why existing prediction systems perform poorly on the VizWiz dataset.  

We also tally how often a visual question is unanswerable, as indicated by at least half the crowdsourced answers for a visual question stating the answer is ``unanswerable" or ``unsuitable image".  We find 28.63\% of visual questions are not answerable.  This finding validates the practical importance of the recent efforts~\cite{KafleKa17,MahendruPrMoBaLe17,RayChBaBaPa16,ToorWeNa17} to augment VQA datasets with irrelevant visual questions.  Moreover, our dataset offers more fine-grained annotations that enable research to automatically identify whether the answerability issue is due to inadequate image quality (e.g., ``Unsuitable Image") or image content (i.e., ``Unanswerable").  

We also analyze answer diversity by computing statistics for the number of words in each answer.  The median and mean answer lengths are 1.0 and 1.66 words respectively.  These statistics resemble what is observed for numerous artificially constructed VQA datasets, as summarized in ~\cite{DasKoGuSiYaMoPaBa17} and ~\cite{KafleKa17}.  We also compute the percentage of answers with different answer lengths: 67.32\% have one word, 20.74\% have two words, 8.24\% have three words, 3.52\% have four words, and the remaining 0.01\% have more than four words.  Interestingly, our answers are longer on average than observed by Antol et al.~\cite{AntolAgLuMiBaZiPa15}, who used a similar crowdsourcing system.  We attribute this discrepancy in part to many VizWiz visual questions asking to read multi-word text.

We finally compute the level of human agreement on answers, using exact string matching.  
Despite that humans provided open-ended text as answers, we observe agreement from independent people on the answer for most visual questions (i.e., 97.7\%).  More than three people agreed on the most popular answer for 72.83\% of visual questions, exactly three people agreed for 15.5\% of visual questions, and exactly two people agreed for 9.67\% of visual questions.  This agreement level is the lower bound since less stringent agreement measures (e.g., that resolve synonyms) may lead to greater agreement. 
  
\section{VizWiz Benchmarking}
We now investigate the difficulty of the VizWiz dataset for existing algorithms.   We divide the final dataset into training, validation, and test sets of 20,000, 3,173, and 8,000 visual questions, respectively (i.e, approximately a 65/10/25 split).  All results below are reported for the test dataset.  

\subsection{Visual Question Answering}
We assess the difficulty of the VizWiz dataset for modern VQA algorithms and evaluate how well models trained on VizWiz generalize (more details in Supp. Materials).

\vspace{-1em}
\paragraph{Baselines.}  We benchmark nine methods.  Included are three top-performing VQA methods~\cite{AndersonHeBuTeJoGoZh17,GoyalKhSuBaPa16,KazemiEl17}, which we refer to as \texttt{Q+I}~\cite{KazemiEl17}, \texttt{Q+I+A}~\cite{GoyalKhSuBaPa16}, and \texttt{Q+I+BUA}~\cite{AndersonHeBuTeJoGoZh17}.  These baselines are trained on the VQA V2.0 dataset~\cite{GoyalKhSuBaPa16} to predict the 3,000 most frequent answers in the training dataset.  ~\cite{GoyalKhSuBaPa16} relies on image and question information alone, ~\cite{KazemiEl17} adds an attention mechanism to specify image regions to focus on, and ~\cite{AndersonHeBuTeJoGoZh17} combines bottom-up and top-down attention mechanisms to focus on objects and other salient image regions.  We introduce three fine-tuned classifiers built on the three networks, which we refer to as \texttt{FT}~\cite{GoyalKhSuBaPa16}, \texttt{FT}~\cite{KazemiEl17}, and \texttt{FT}~\cite{AndersonHeBuTeJoGoZh17}. We also train the three networks from scratch using the VizWiz data alone, and we refer to these as \texttt{VizWiz}~\cite{GoyalKhSuBaPa16}, \texttt{VizWiz}~\cite{KazemiEl17}, and \texttt{VizWiz}~\cite{AndersonHeBuTeJoGoZh17}.

\vspace{-1em}
\paragraph{Evaluation Metrics.}
We evaluate with respect to four metrics: Accuracy~\cite{AntolAgLuMiBaZiPa15}, CIDEr~\cite{VedantamZiPa15}, BLEU4~\cite{PapineniRoWaZh02}, and METEOR~\cite{ElliottKe13}.  Accuracy~\cite{AntolAgLuMiBaZiPa15} was introduced as a good metric when most answers are one word.  Since nearly half the answers in VizWiz exceed one word, we also use image description metrics provided by \cite{ChenFaLiVeGuArZi15} which are designed for evaluating longer phrases and/or sentences.

\begin{table}[b!]
  \centering
        \begin{tabular}{ l  l  l  l l  l }
    \toprule
     Method  & Acc & CIDEr & BLEU & METEOR \\ 
    \midrule
       \textbf{Q+I~\cite{GoyalKhSuBaPa16}} & 0.137 & 0.224 & 0.000 & 0.078 \\
       \textbf{Q+I+A~\cite{KazemiEl17}} & 0.145 & 0.237 & 0.000 & 0.082   \\
       \textbf{Q+I+BUA~\cite{AndersonHeBuTeJoGoZh17}} & 0.134 & 0.226	
 & 0.000 & 0.077 	
   \\
       \textbf{FT~\cite{GoyalKhSuBaPa16}} & 0.466 & 0.675 & 0.314 & 0.297  \\
       \textbf{FT~\cite{KazemiEl17}} & 0.469 & 0.691 & 0.351 & 0.299 \\
       \textbf{FT~\cite{AndersonHeBuTeJoGoZh17}} & \textbf{0.475} & \textbf{0.713} & 0.359 & \textbf{0.309}   \\
       \textbf{VizWiz~\cite{GoyalKhSuBaPa16}} & 0.465 & 0.654 & 0.353 & 0.298  \\
       \textbf{VizWiz~\cite{KazemiEl17}} & 0.469 & 0.661 & 0.356 & 0.302 \\
       \textbf{VizWiz~\cite{AndersonHeBuTeJoGoZh17}} & 0.469 & 0.675 & \textbf{0.396} & 0.306   \\
       \bottomrule	 
  \end{tabular}
          \vspace{0.5em}
        \caption{Performance of VQA methods on the VizWiz test data with respect to four metrics.  Results are shown for three variants of three methods~\cite{AndersonHeBuTeJoGoZh17,GoyalKhSuBaPa16,KazemiEl17}: use models as is, fine-tuned (FT), and trained on only VizWiz data (VizWiz).  The methods use different combinations of image (I), question (Q), and attention (A) models.}  
        \label{table_vqaPerformance}
\end{table} 

 \begin{table}[h!]
  \centering
        \begin{tabular}{ l  l  l  l l  l  }
    \toprule
       & Yes/No & Number & Unans & Other \\ 
    \midrule
       \textbf{Q+I~\cite{GoyalKhSuBaPa16}} & 0.598 & 0.045 & 0.070 & 0.142  \\
       \textbf{Q+I+A~\cite{KazemiEl17}} & 0.605 & 0.068 & 0.071 & 0.155  \\
       \textbf{Q+I+BUA~\cite{AndersonHeBuTeJoGoZh17}} & 0.582 & 0.071 & 0.060 & 0.143  \\
       \textbf{FT~\cite{GoyalKhSuBaPa16}} & 0.675 & 0.220 & 0.781 & 0.275 \\
       \textbf{FT~\cite{KazemiEl17}} & \textbf{0.681} & 0.213 & 0.770 & 0.287  \\
       \textbf{FT~\cite{AndersonHeBuTeJoGoZh17}} & 0.669 & 0.220 & 0.776 & \textbf{0.294}  \\
       \textbf{VizWiz~\cite{GoyalKhSuBaPa16}} & 0.597 & \textbf{0.262} & \textbf{0.805} & 0.264  \\
       \textbf{VizWiz~\cite{KazemiEl17}} & 0.608 & 0.218 & 0.802 & 0.274  \\
       \textbf{VizWiz~\cite{AndersonHeBuTeJoGoZh17}} & 0.596 & 0.210 & \textbf{0.805} & 0.273  \\
       \bottomrule	 
  \end{tabular}
  \vspace{0.25em}
        \caption{Accuracy of nine VQA algorithms for visual questions that lead to different answer types.}  
  \vspace{-0.5em}
        \label{table_fineGrainedVqaPerformance}
\end{table}

\vspace{-1em}
\paragraph{Results.}
We first analyze how existing prediction models~\cite{AndersonHeBuTeJoGoZh17,GoyalKhSuBaPa16,KazemiEl17} perform on the VizWiz test set.  As observed in the first three rows of \textbf{Table~\ref{table_vqaPerformance}}, these models perform poorly, as indicated by low values for all metrics; e.g., $\sim$0.14 accuracy for all algorithms.  We attribute the poor generalization of these algorithms largely to their inability to predict answers observed in the VizWiz dataset; i.e., only 824 out of the top 3,000 answers in VizWiz are included in the dataset (i.e., VQA 2.0~\cite{GoyalKhSuBaPa16}) used to train the models.

We observe in \textbf{Table~\ref{table_vqaPerformance}} that fine-tuning (i.e., rows 4--6) and training from scratch (i.e., rows 7--9) yield significant performance improvements over relying on the three prediction models~\cite{AndersonHeBuTeJoGoZh17,GoyalKhSuBaPa16,KazemiEl17} as is.  We find little performance difference between fine-tuning and training from scratch for the three models.  While the number of training examples in VizWiz is relatively small, we hypothesize the size is sufficient for teaching the models to retain knowledge about answer categories that are applicable in this setting.  Despite the improvements, further work is still needed to achieve human performance (i.e., 0.75 accuracy)\footnote{Performance is measured by partitioning the dataset into 10 sets of one answer per visual question and then evaluating one answer set against the remaining nine answer sets for all 10 partitions using the accuracy metric.}. 

We next analyze what predictive cues may lead to algorithm success/failure.  We observe models that add the attention mechanism~\cite{AndersonHeBuTeJoGoZh17,KazemiEl17} consistently outperform relying on image and question information alone~\cite{GoyalKhSuBaPa16}.  Still, the improvements are relatively small compared to improvements typically observed on VQA datasets.  We hypothesize this improvement is small in part because many images in VizWiz include few objects and so do not need to attend to specific image regions.  We also suspect attention models perform poorly on images coming from blind photographers since such models were not trained on such images.

We further enrich our analysis by evaluating the nine algorithms for visual questions that lead to different answer types (their frequencies in VizWiz are shown in parentheses): ``yes/no" (4.80\%), ``number" (1.69\%), ``other" (58.91\%), and ``unanswerable" (34.6\%).  Results are shown in \textbf{Table~\ref{table_fineGrainedVqaPerformance}}.  Overall, we observe performance gains by fine-tuning algorithms (rows 4--6) and training from scratch (rows 7--9), with the greatest gains for ``unanswerable" visual questions and smallest gains for ``number" and ``other" visual questions.  Exemplar failures include when asking for text to be read (e.g., captchas, cooking directions) and things to be described (e.g., clothes).

Finally, we evaluate how well algorithms trained on VizWiz predict answers for the VQA 2.0 test dataset~\cite{GoyalKhSuBaPa16}.  The six models that are fine-tuned and trained from scratch for the three models~\cite{AndersonHeBuTeJoGoZh17,GoyalKhSuBaPa16,KazemiEl17} do not generalize well; i.e., accuracy scores range from 0.218 to 0.318.  This result suggests that VizWiz provides a domain shift to a different, difficult VQA environment compared to existing datasets.  

\subsection{Visual Question Answerability}
We next turn to the question of how accurately an algorithm can classify a visual question as answerable.  

\vspace{-1em}
\paragraph{Baselines.}
We benchmark eight methods.  We use the only publicly-available method for predicting when a question is not relevant for an image~\cite{MahendruPrMoBaLe17}.
This method uses NeuralTalk2~\cite{KarpathyFe15} pre-trained on the MSCOCO captions dataset~\cite{LinMaBeHaPeRaDoZi14} to generate a caption for each image.  The algorithm then measures the similarity between the proposed caption and the question to predict a relevance score.  The model is trained on the QRPE dataset~\cite{MahendruPrMoBaLe17}.  We use the model as is (i.e., \texttt{Q+C}~\cite{MahendruPrMoBaLe17}), fine-tuned to the VizWiz data (i.e., \texttt{FT}~\cite{MahendruPrMoBaLe17}), and trained from scratch on the VizWiz data only (i.e., \texttt{VizWiz}~\cite{MahendruPrMoBaLe17}).  We also employ our top-performing VQA algorithm by using its output probability that the predicted answer is ``unanswerable" (\texttt{VQA}~\cite{GoyalKhSuBaPa16}).  We enrich our analysis by further investigating the influence of different features on the predictions: question alone (i.e., \texttt{Q}), caption alone (i.e., \texttt{C}), image alone using ResNet-152 CNN features (i.e., \texttt{I}), and the question with image (i.e., \texttt{Q+I}).

\vspace{-1em}
\paragraph{Evaluation Metrics.}
We report the performance of each method to predict if a visual question is not answerable using a precision-recall curve.  We also report the average precision (AP); i.e., area under a precision-recall curve.  

\begin{figure}[t]
\centering
\includegraphics[width=0.47\textwidth]{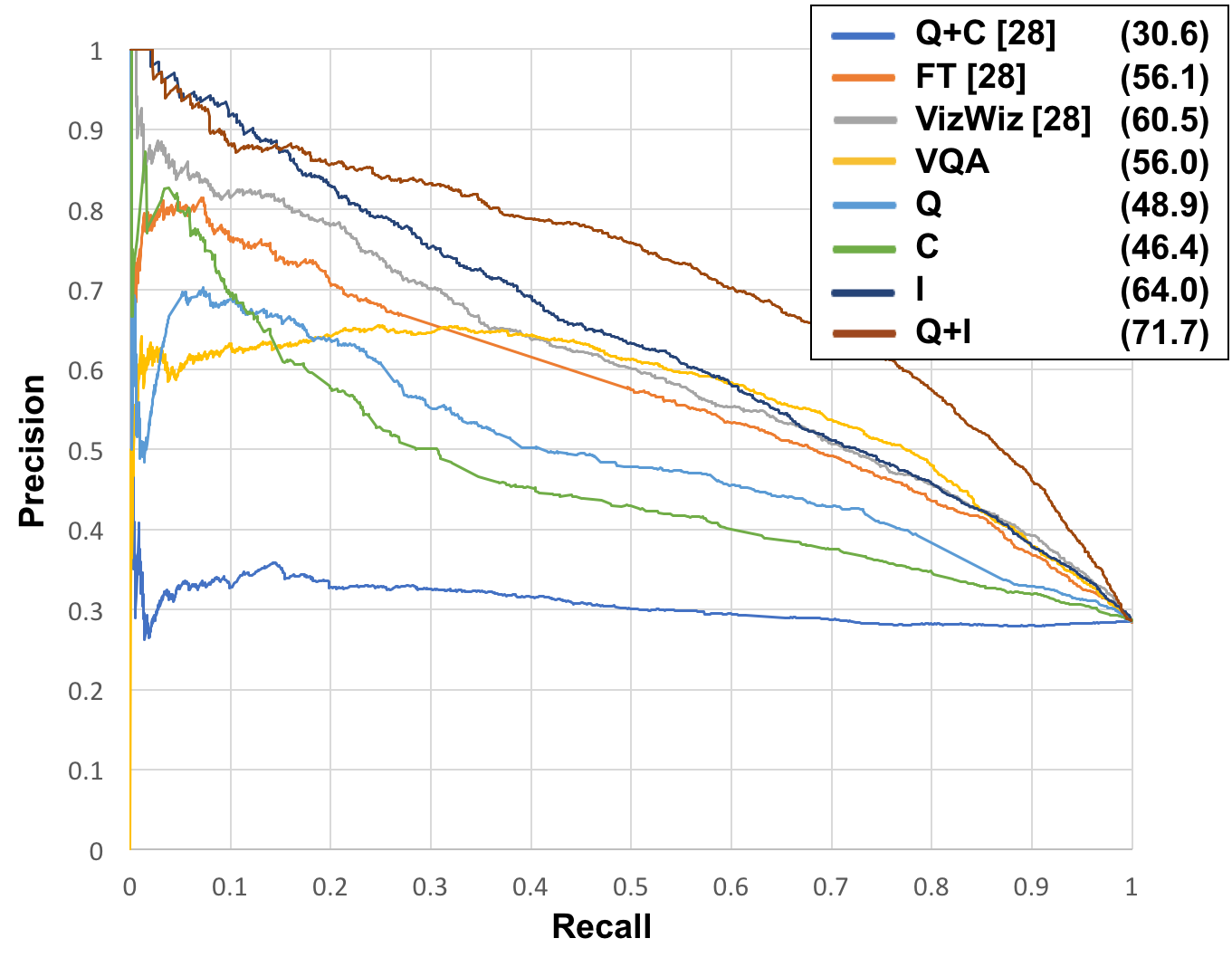}
\caption{Precision-recall curves and average precision scores for the answerability models tested on the VizWiz test dataset.}
\label{fig_answerabilityPRcurves}
\end{figure}

\vspace{-1em}
\paragraph{Results.}
\textbf{Figure~\ref{fig_answerabilityPRcurves}} shows the precision-recall curves.  As observed, all methods outperform the status quo approach by 25\% to 41\%; i.e., AP score of 30.6 for \cite{MahendruPrMoBaLe17} versus 71.7 for \texttt{Q+I}.  We hypothesize this large discrepancy arises because the irrelevance between a question and image arises for more reasons in VizWiz than for QRPE; e.g., low quality images and fingers blocking the camera view.  When comparing the predictive features, we find the image provides the greatest predictive power (i.e., AP = 64) and is solidly improved by adding the question information (i.e., AP = 71.7).  Again, we attribute this finding to low quality images often leading visual questions to be unanswerable.

\section{Conclusions}
We introduced VizWiz, a VQA dataset which originates from a natural use case where blind people took images and then asked questions about them.  Our analysis demonstrates this dataset is difficult for modern algorithms.  Improving algorithms on VizWiz can simultaneously educate people about the technological needs of blind people while providing an exciting new opportunity for researchers to develop assistive technologies that eliminate accessibility barriers for blind people.  We share the dataset and code to facilitate future work (\texttt{http://vizwiz.org/data/}).
  \\

\noindent
\textbf{Acknowledgements}:
We thank Erin Brady, Samuel White, and Yu Zhong for supporting the VizWiz deployment, the anonymous users of the VizWiz application for sharing their visual questions, the authors of \cite{AntolAgLuMiBaZiPa15} for sharing their answer collection code, and the anonymous crowd workers for providing the annotations.  This work is supported by NSF awards to D.G. (IIS-1755593), K.G. (IIS-1514118), and J.L. (IIS-1704309); AWS Machine Learning Research Award to K.G.; Google Research Award to J.P.B.; Microsoft Research support to J.P.B.; and New York State support from the Goergen Institute for Data Science to J.L.

\balance{}
{\small
\bibliographystyle{ieee}
\bibliography{myReferences}
}

\pagebreak
\subsection*{Supplementary Materials}
This document supplements the main paper with the following details about:

\begin{enumerate}
\item I - Filtering visual questions (supplements \textbf{Section 3.2}).
\item II - Collecting answers to visual questions (supplements \textbf{Section 3.3}).
\item III - Analyzing the VizWiz dataset (supplements \textbf{Section 4}).
\item IV - Benchmarking algorithm performance (supplements \textbf{Section 5}).
\end{enumerate}

\section*{I - Filtering Visual Questions}
\begin{figure*}[h!]
\centering
\includegraphics[width=0.65\textwidth]{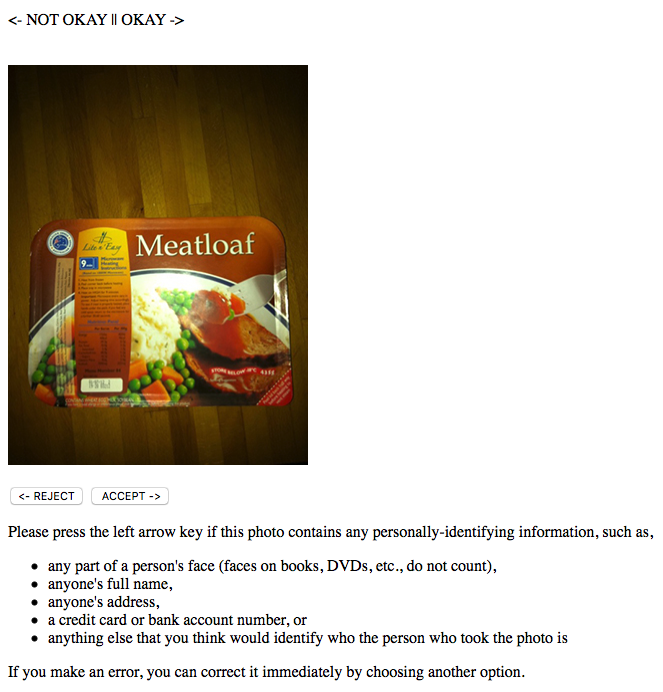}
\caption{AMT user interface for identifying images showing PII.}
\label{fig_crowdsourcingPII}
\end{figure*}

We first used the crowdsourcing system shown in \textbf{Figure~\ref{fig_crowdsourcingPII}} to identify images showing personal-identifying information.  To err on the safe side in protecting all involved parties, we next iteratively developed a taxonomy of possible vulnerabilities people face when working with a VQA dataset created ``in the wild".  During an initial brainstorming session, we identified the following three categories: (1) personally-identifying information, also called PII (e.g., any part of a person's face, financial statements, prescriptions), (2) Location (e.g., addressed mail, business locations), and (3) Adult Content (e.g., nudity, cuss words).  We then examined the robust-ness of this taxonomy by evaluating the inter-annotator agreement between three domain experts who reviewed 1,000 randomly-selected visual questions and labeled ``vulnerable" instances.  We found exactly one person marked a visual question for removal for the majority of instances (i.e., 44) that visual questions were tagged for removal (i.e., 64).  We found most disagreements occurred on visual questions for which the researchers were not sure, such as in poor quality images or complex scenes.  We therefore added two more categories to our taxonomy that reflected our desire to err on the safe side: (4) Questionable Complex Scenes and (5) Questionable Low Quality Images.  

\section*{II - Answer Collection}
\subsection*{Answer Post-Processing}
Following prior work~\cite{AntolAgLuMiBaZiPa15},  we converted all letters to lower case, converting numbers to digits, and removing punctuation and articles (i.e., ``a", ``an", ``the").  We further post-processed the answers by fixing spelling mistakes and removing filler words (i.e., ``it'", ``is", ``its", ``and", ``\&", ``with", ``there", ``are", ``of", ``or").  For spell checking, we relied on two automated spell-checkers to reveal which words in the answers neither reflected common nor popular modern words: (1) Enchant\footnote{\texttt{https://www.abisource.com/projects/enchant/}} provides an API to multiple libraries such as Aspell/Pspell and AppleSpell and (2) an algorithm invented by Google search quality director Peter Norvig\footnote{\texttt{http://norvig.com/spell-correct.html}}, that is based on frequent words in popular Wikipedia articles and movie subtitles, and so augments modern words such as iPhone and Gmail.  Both the aforementioned tools also employ different mechanisms to return correct word candidates.  When the most probable correct word from both tools matched, we replaced the original word with the candidate.  For the remaining answers, we solicited the correct spelling of the word from trusted in-house human reviewers.  We found many of the detected ``misspelled" words were valid captchas and so did not need spell-correction.  

\subsection*{Crowdsourcing System}
\begin{figure*}[t!]
\centering
\includegraphics[width=0.96\textwidth]{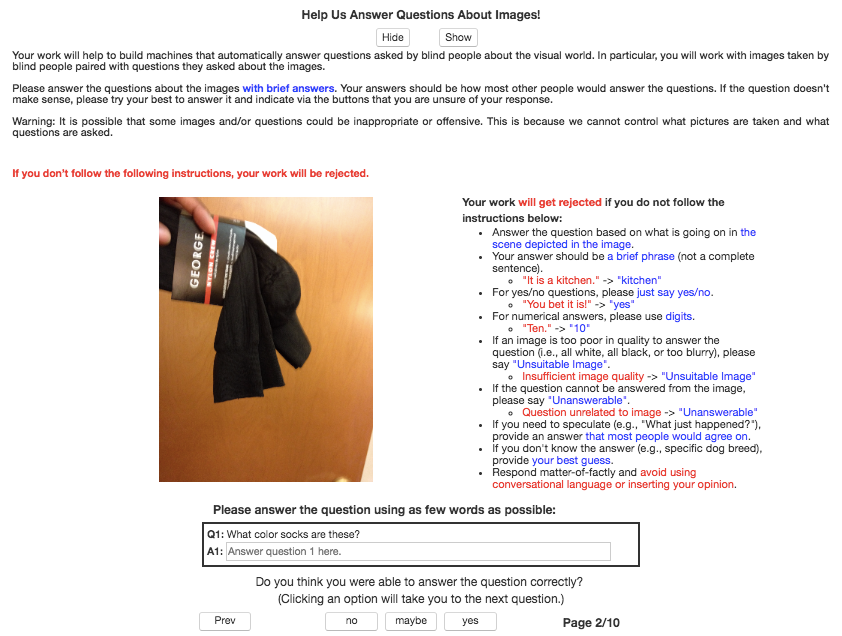}
\caption{AMT user interface for collecting answers to visual questions.}
\label{fig_crowdsourcingAnswers}
\end{figure*}

We show the Amazon Mechanical Turk (AMT) interface that we used to collect answers in \textbf{Figure~\ref{fig_crowdsourcingAnswers}}.  We limited our users to US citizens to minimize concerns about whether a person is familiar with the language.  We also limited our users to those who previously had 95\% jobs approved for over 500 jobs to increase the likelihood of collecting high quality results.  Finally, we used the ``Adult Qualification" in AMT to ensure our selected crowd was comfortable reviewing adult content.  This was important because visual questions are gathered ``from the wild" so could contain content that is not appropriate for a general audience (e.g., nudity).

\section*{III - VQA Dataset Analysis}
\subsection*{Question Length Distribution}
\begin{figure*}[t!]
\centering
\includegraphics[width=0.65\textwidth]{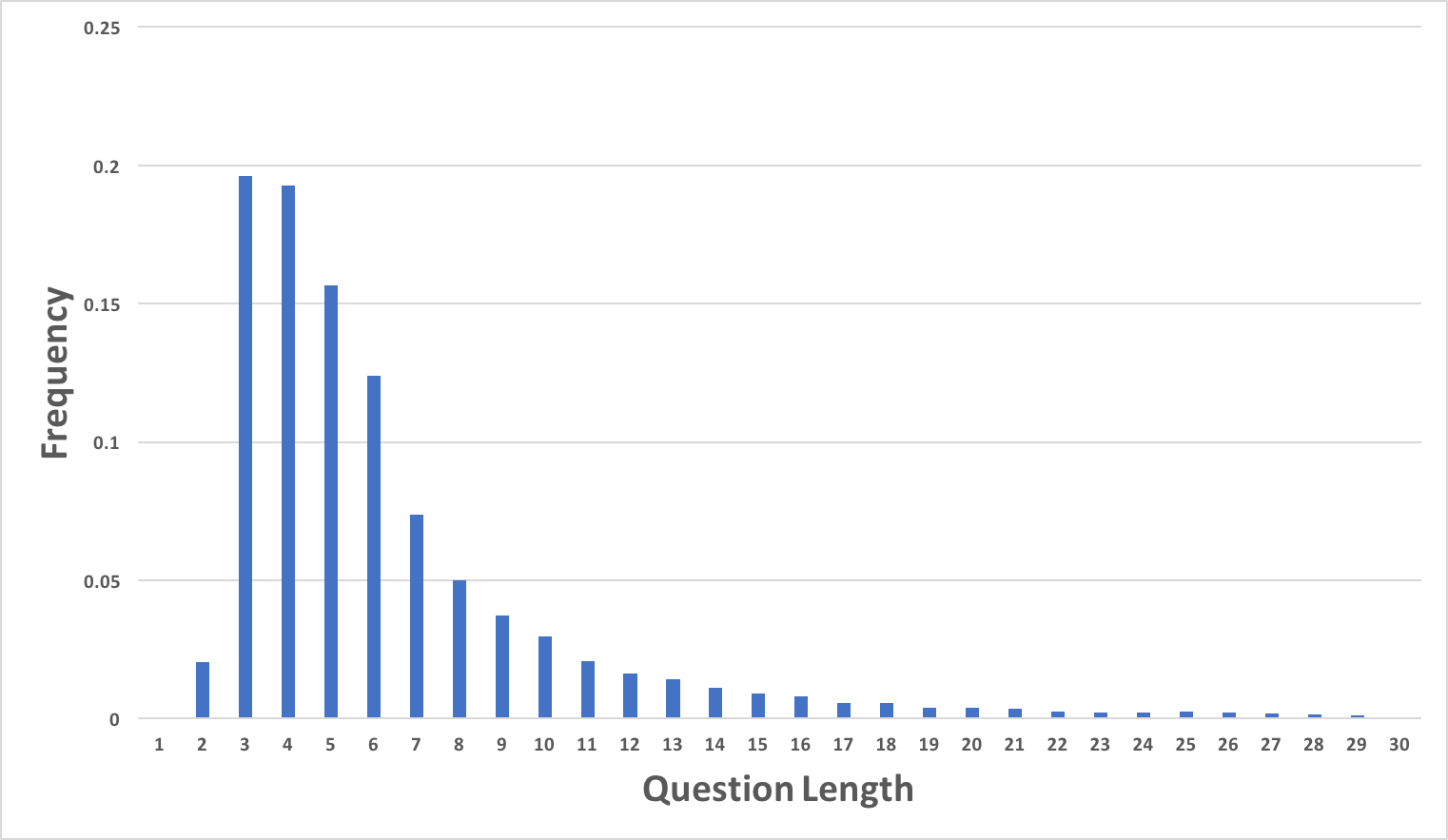}
\caption{Distribution of number of words per visual question.}
\label{fig_questionLengths}
\end{figure*}

We augment the statistics supplied in the main paper, with the fine-grained distribution showing the number of words in each visual question in \textbf{Figure~\ref{fig_questionLengths}}.  We cut the plot off at 30 words in the visual question\footnote{There is a small tail of visual questions that spread to a maximum of 62 words in the question.}.  This distribution highlights the prevalence of outliers with few words or 10s of words in the question.

\begin{figure}[t!]
\centering
\includegraphics[width=0.1\textwidth]{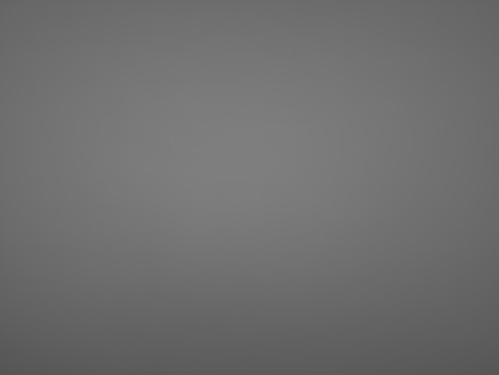}
\caption{The average image created using all images in VizWiz, excluding those that are in unanswerable visual questions.}
\label{fig_averageImageExcludeUnanswerables}
\end{figure}

\begin{figure}[t!]
\centering
\begin{subfigure}{.48\textwidth}
  \centering
  \includegraphics[width=1\linewidth]{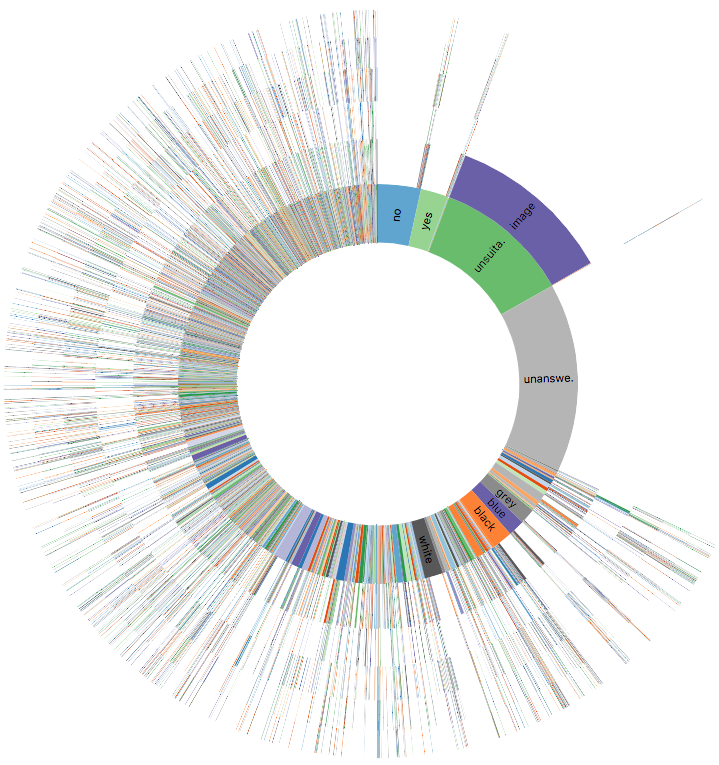}
    \caption{}
\end{subfigure}
\begin{subfigure}{.48\textwidth}
  \centering
  \includegraphics[width=1\linewidth]{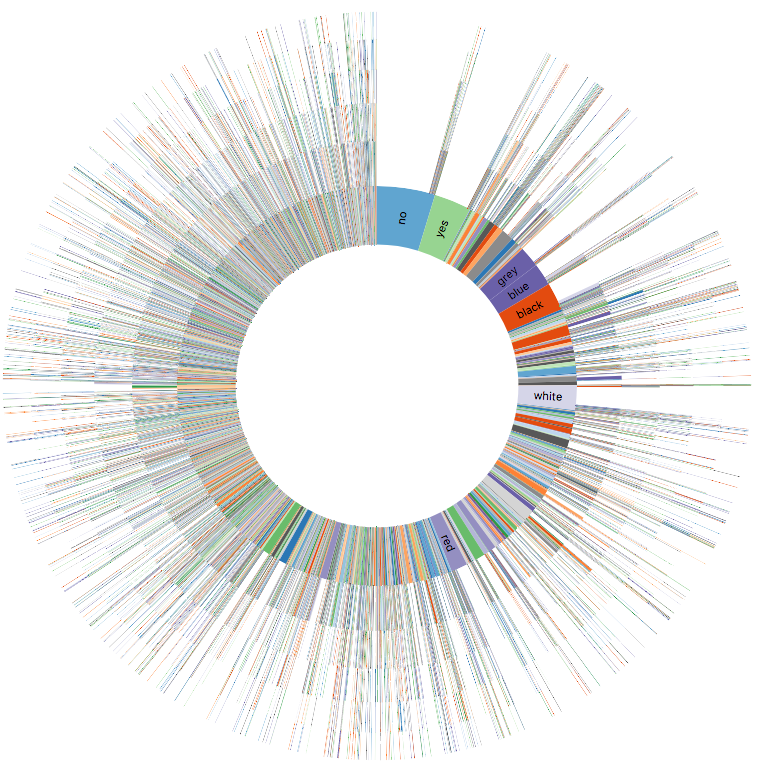}
  \caption{}
\end{subfigure}
\caption{Distribution of the first six words for (a) all answers in VizWiz and (b) all answers in VizWiz excluding unanswerable visual questions.  The innermost ring represents the first word and each subsequent ring represents a subsequent word.  The arc size is proportional to the number of answers with that initial word/phrase.}
\label{fig_answerSunburstDiagram}
\end{figure}

\subsection*{Average Image Excluding ``Unanswerable" Visual Questions}
We show a parallel image supplied in the main paper here, with the only change being that we show the average of all images excluding those coming from visual questions labelled as unanswerable.  The resulting image shown in \textbf{Figure~\ref{fig_averageImageExcludeUnanswerables}} resembles that shown in the main paper by also being a gray image, and so reflecting a diverse set of images that do not conform to a particular structure. 

\subsection*{Answer Analysis}
We show in \textbf{Figure~\ref{fig_answerSunburstDiagram}} sunburst diagrams which visualize the frequency that answers begin with different words/phrases.  The most common answers, following ``Unsuitable Image" and ``Unanswerable", are yes, no, and colors.  We observe there is a large diversity of uncommon answers as well as answer lengths spanning up to 6 words long.  

\begin{figure*}[t!]
\centering
\begin{subfigure}{.6\textwidth}
  \centering
  \includegraphics[width=1\linewidth]{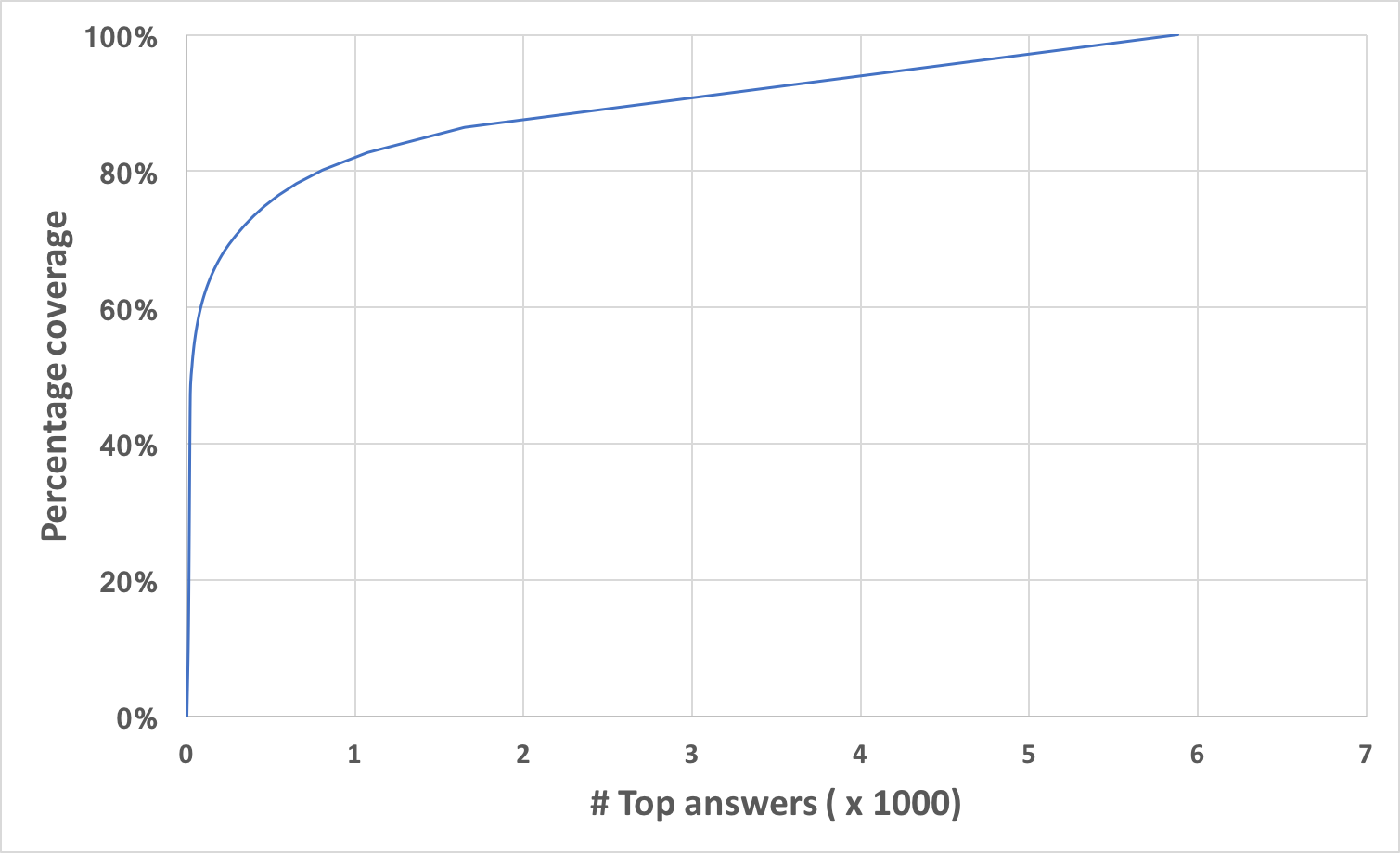}
    \caption{}
\end{subfigure}
\begin{subfigure}{.6\textwidth}
  \centering
  \includegraphics[width=1\linewidth]{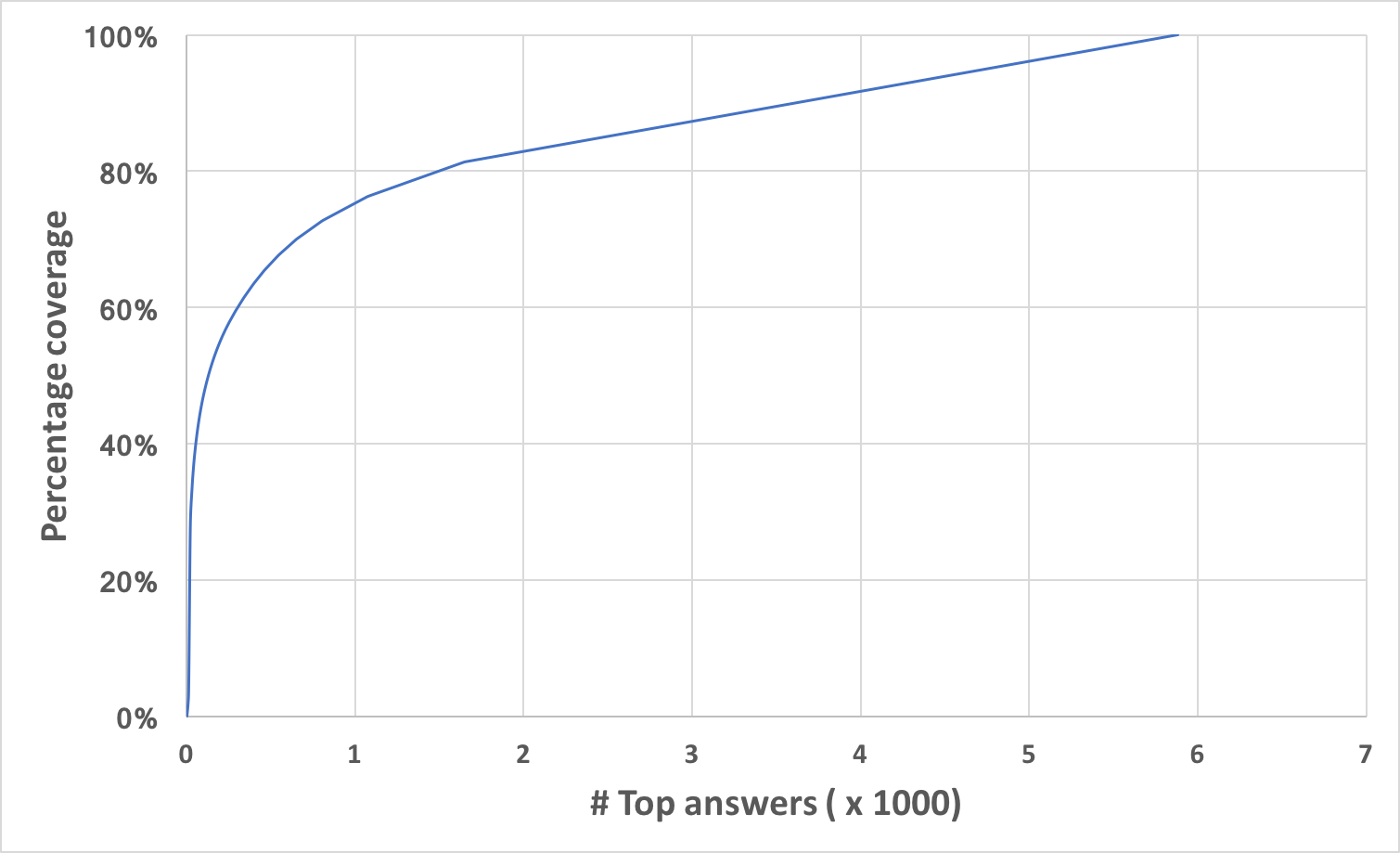}
  \caption{}
\end{subfigure}
\caption{Cumulative number of visual questions covered by the most frequent answers in VizWiz for (a) all answers in VizWiz and (b) all answers in VizWiz excluding unanswerable visual questions.}
\label{fig_cumulativeAnswerCoverage}
\end{figure*}

We also show in \textbf{Figure~\ref{fig_cumulativeAnswerCoverage}} plots of the cumulative coverage of all answers versus the most frequent answers.  The straight line with a slope of roughly 1 further illustrates the prevalence of a long tail of unique answers.  

\section*{IV - VizWiz Algorithm Benchmarking}
\subsection*{VQA}
In the main paper, we report results for fine-tuned models.  We fine-tune each pre-trained model on VizWiz using the most frequent 3,000 answers in the training set of VizWiz.  For the initialization of the last layer, if the answer is in the candidate answer set of VQA V2.0 dataset~\cite{GoyalKhSuBaPa16}, we initialize the corresponding parameters from the pre-trained model, and if not, we randomly initialize the parameters. We use Adam solver~\cite{KingmaBa14} with a batch size of 128 and an initial learning rate of 0.01 that is dropped to 0.001 after the first 10 epochs.  The training is stopped after another 10 epochs. We employ both dropout~\cite{SrivastavaHiKrSuSa14} and batch normalization~\cite{IoffeSz15} during training.  
  
In the main paper, we also report results for models trained from scratch.  Each model is trained using the 3,000 most frequent answers in the train split of VizWiz.  We initialize all parameters in the model to random values.

Finally, we report fine-grained details to expand on our findings reported in the main paper about how well algorithms trained on VizWiz predict answers for the VQA 2.0 test dataset~\cite{GoyalKhSuBaPa16}.  We report results for the six models that are fine-tuned and trained from scratch for the three models~\cite{AndersonHeBuTeJoGoZh17,GoyalKhSuBaPa16,KazemiEl17} with respect to all visual questions as well as with respect to the four answer types in \textbf{Table~\ref{table_crossDatasetPerformance}}.  These results highlight that VizWiz provides a domain shift to a different, difficult VQA environment compared to existing datasets.  

\begin{table}[h!]
  \centering
        \begin{tabular}{  l  l  l  l  l  }
    \toprule
       &  All & Yes/No & Number & Other \\ 
    \midrule
       \textbf{FT~\cite{GoyalKhSuBaPa16}} & 0.300 & 0.612 & 0.094 &  0.079  \\
       \textbf{FT~\cite{KazemiEl17}} & 0.318 & 0.601 & 0.163 & 0.110 \\
       \textbf{FT~\cite{AndersonHeBuTeJoGoZh17}} & 0.304 & 0.595 & 0.082 & 0.105 \\
       \textbf{VizWiz~\cite{GoyalKhSuBaPa16}} & 0.218 & 0.461 & 0.074 & 0.042 \\
       \textbf{VizWiz~\cite{KazemiEl17}} & 0.228 & 0.465 & 0.131 & 0.049 \\
       \textbf{VizWiz~\cite{AndersonHeBuTeJoGoZh17}} & 0.219 & 0.453 & 0.083 & 0.048 \\
       \bottomrule	 
  \end{tabular}
  \vspace{0.25em}
        \caption{Shown is the cross-dataset performance of six models trained on VizWiz and tested on the VQA 2.0 test dataset~\cite{GoyalKhSuBaPa16}.}  
  \vspace{-0.5em}
        \label{table_crossDatasetPerformance}
\end{table} 

\subsection*{Answerability}
Below is a brief description of the implementations of the models we use in the main paper:

\begin{itemize}
\item \texttt{Q}: a one-layer LSTM is used to encode the question and is input to a softmax layer.
\item \texttt{C}: a one-layer LSTM is used to encode the caption and is input to a softmax layer.
\item \texttt{I}: ResNet-152 is used to extract the image features from the pool5 layer and is input to a softmax layer.
\item \texttt{Q+C}: the question and caption are encoded by two separate LSTMs and then the features of the question and caption are concatenated and input to a softmax layer.
\item \texttt{Q+I} the features of question and image are concatenated and input to a softmax layer.
\end{itemize}

For the fine-tuned model, we initialize the parameters using the pre-trained model.  We train from scratch by randomly initializing the parameters.  For both approaches, we train for 10 epochs on the VizWiz dataset.

We augment here our findings of the average precision in the main paper with the average F1 score in \textbf{Table~\ref{table_APandF1}}.  As observed, the top-performing method remains \texttt{Q+I} whether using the AP score or F1 score.  

\begin{table}[h!]
  \centering
        \begin{tabular}{  l  l  l  }
    \toprule
       Model &  Average Precision & Average F1 score \\ 
    \midrule
       \texttt{Q+C}~\cite{MahendruPrMoBaLe17} & 0.306 & 0.383 \\
       \texttt{FT}~\cite{MahendruPrMoBaLe17} & 0.561 & 0.542 \\
       \texttt{VizWiz}~\cite{MahendruPrMoBaLe17} & 0.605 & 0.549 \\
       \texttt{VQA}~\cite{GoyalKhSuBaPa16} & 0.560 & 0.569 \\
       \texttt{Q} & 0.490 & 0.233 \\
       \texttt{C} & 0.464 & 0.270 \\
       \texttt{I} & 0.640 & 0.518 \\
       \texttt{Q+I} & 0.717 & 0.648 \\
       \bottomrule	 
  \end{tabular}
  \vspace{0.25em}
        \caption{Shown are the average precision scores and average F1 scores for eight models used to predict whether a visual question is answerable.}  
  \vspace{-0.5em}
        \label{table_APandF1}
\end{table} 

\begin{figure*}[t!]
\centering
\begin{subfigure}{1\textwidth}
  \centering
  \includegraphics[width=1\linewidth]{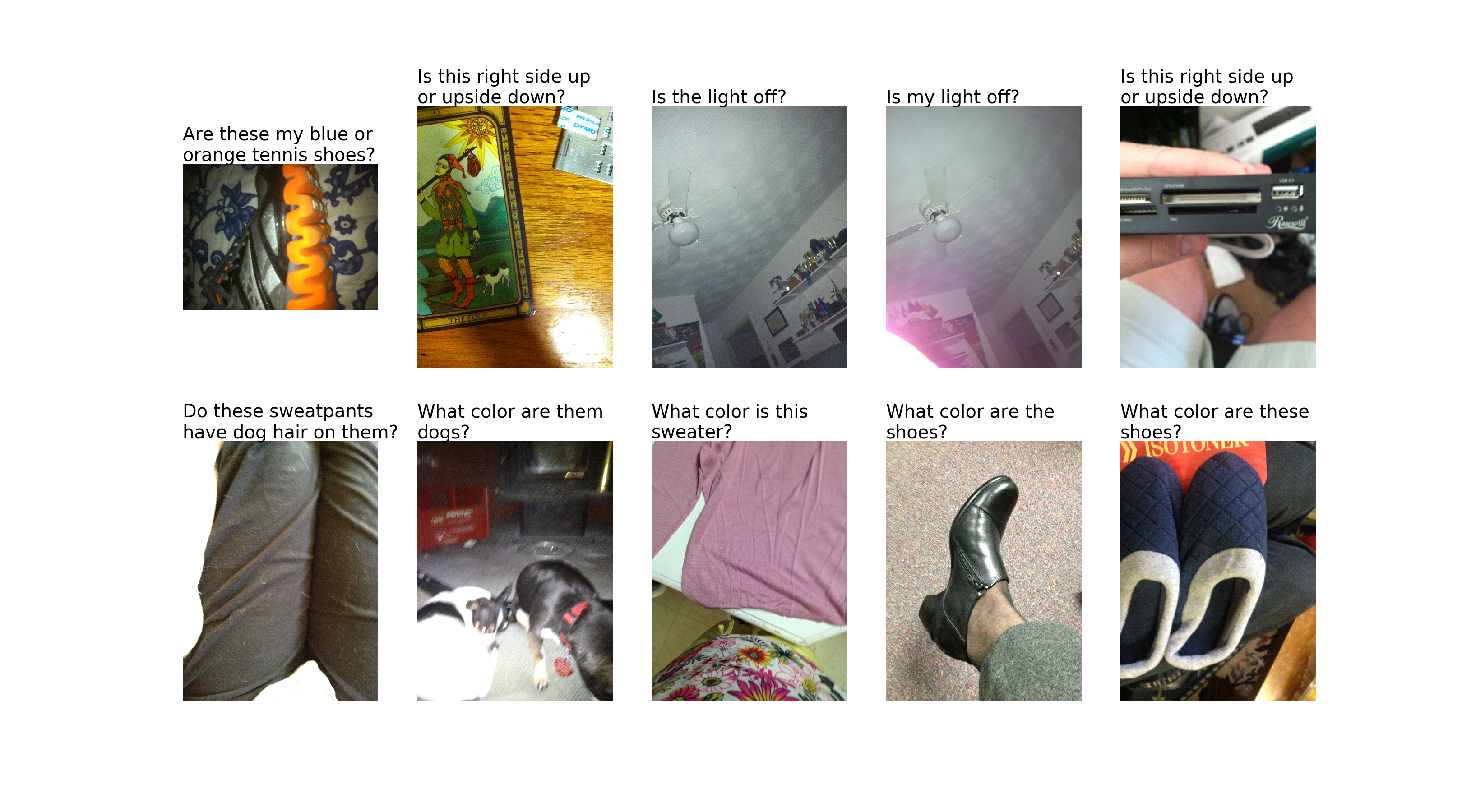}
  \vspace{-5em}
    \caption{}
\end{subfigure}
\begin{subfigure}{1\textwidth}
  \centering
  \includegraphics[width=1\linewidth]{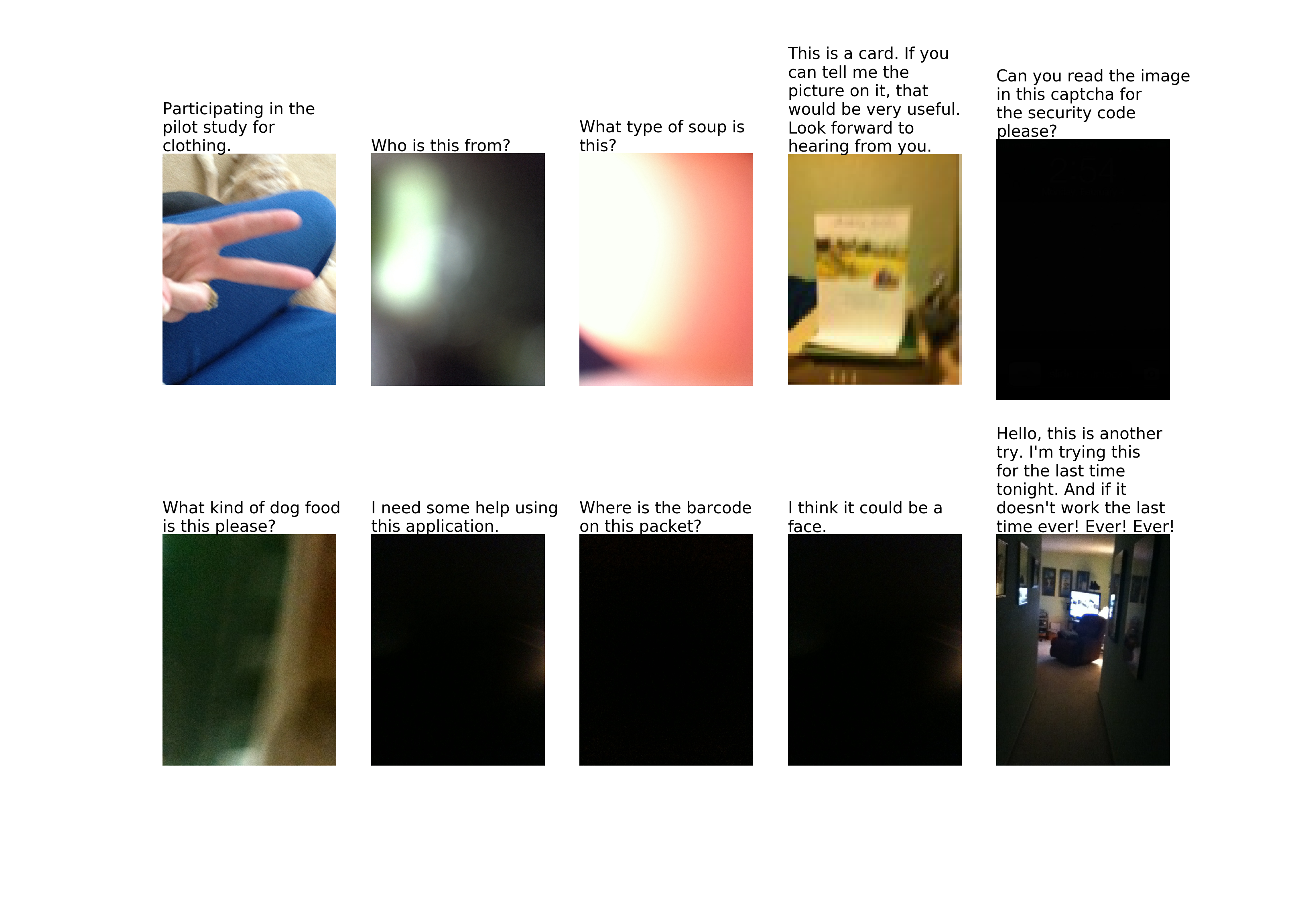}
  \vspace{-7.2em}
  \caption{}
\end{subfigure}
\caption{Top 10 most confident predictions by the top-performing \texttt{Q+I} model for visual questions in the VizWiz test dataset that are (a) answerable and (b) unanswerable.}
\label{fig_answerablePredictions}
\end{figure*}

We also show the top 10 most confident answerable and answerable predictions for the top-performing \texttt{Q+I} implementation  in \textbf{Figure~\ref{fig_answerablePredictions}}.  Our findings highlight how predictive cues may relate to the quality of images and specific questions (e.g., ``What color...?"). 

\end{document}